# Contrastive Learning and the Emergence of Attributes Associations

Daniel N. Nissani (Nissensohn)[1]

[1]Independent Research

dnissani@post.bgu.ac.il

**Abstract.** In response to an object presentation, supervised learning schemes generally respond with a parsimonious label. Upon a similar presentation we humans respond again with a label, but are flooded, in addition, by a myriad of associations. A significant portion of these consist of the presented object attributes (e.g. "comprises one long quasi vertical segment"). Contrastive learning is a self supervised learning scheme based on the application of identity preserving transformations on the object input representations. It is intuitively assumed in this work that the same applied transformations preserve, in addition to the identity of the presented object, also the identity of its semantically meaningful attributes. The natural outcome of this intuition is that the output representations of such a contrastive learning scheme contain valuable information not only for the classification of the presented object, but also for the decision regarding the presence or absence of any attribute of interest. Simulation results which demonstrate this idea and the feasibility of this hypothesis are presented.

**Keywords:** Contrastive Learning, Self-supervised Learning, Learning Representations, Relations, Attributes

## 1    Introduction

It is our goal in this paper to point out to a computational scheme by which attribute relations naturally and seamlessly emerge as result of sensory perception.

In response to an image presentation a supervised trained neural network such as ResNet (He et al., 2015) outputs a parsimonious label (e.g. "horse"). Unless explicitly supervised trained for another purpose (e.g. Rumelhart, 1990), no additional information is usually extracted from the presented stimulus.

In contrast, and as we know by introspection, when presented with a familiar image, we humans respond again with a label, but are flooded, in addition, by a myriad of associations. These can be divided into extrinsic and intrinsic relations. The extrinsic relations can be expressed as triplets relating the subject image to external objects or, more generally, to the environment ("horse eats grass"). The intrinsic or attribute relations, on the other hand, relate the subject image to its intrinsic properties or attributes ("horse has elongated face").

It is these fore mentioned intrinsic attribute relations which are the subject of this paper.

Within the field of artificial neural networks it is useful and common practice to divide the task of classification into two stages, namely 1. the non-linear mapping from raw input (say pixel level) representations to linearly separable output representations (by 'linearly separable output representations' we more precisely mean that the set of representations created by samples of a given class can be separated, fully or mostly, from all other samples by an hyperplane); and 2. the manipulation of this fore mentioned hyperplane into its optimal position, by means of training a supervised or unsupervised linear classifier (Zhang et al., 2016; Nissani (Nissensohn), 2018, respectively).

Contrastive learning is a self-supervised learning scheme which has attracted significant attention in recent years. It has achieved classification results which compete well with those of state of the art supervised learning schemes. In a typical implementation (e.g. Chen et al., 2020) it applies 2 'benign' transformations to the raw (input) representation of each sample in a training mini-batch. This is followed by minimizing the distance between both neural network output representations of this transformed sample while simultaneously maximizing the sum of the distances between the output representations of one of these transformed samples and all the other transformed samples in a mini-batch. By benign transformations we mean such that preserve the identity of the perceived object, that is the resulting transformed raw representation can still be visually recognized as belonging to the same class. Permutating the pixels of an image is *not* a benign transformation, while translating, resizing, or mildly deforming or rotating the image is: its associated class can be still clearly recognized. Amongst all benign transformations, the most efficient for contrastive learning purpose are those which when applied to any sample of a given class distribution result in another sample belonging to the same class distribution. The implementation outlined above tacitly assumes that all samples for which distances are maximized belong to classes different than that for which distance is minimized: this seems to be a valid approximation, especially in the large number of classes' scenario.

After contrastive training completion the resultant output representations (as typically captured from a near to last layer of the network) have been shown to exhibit good linear separability and the application of a (supervised) linear classifier to these representations has resulted in competitive classification accuracy (Chen et al., 2020).

At a bird's eye view the core idea behind contrastive learning may be interpreted as the exploitation of symmetry transformations under which the identity or class of a studied object is maintained invariant. Numerous important laws of physics have resulted during the last few centuries from application of symmetry transformations under which some entity (such as the Lagrangian) is maintained invariant and it is thus not surprising that similar principles may successfully apply to pattern recognition as well.

In Section II we present our main hypothesis regarding the natural emergence of attribute relations. Section III demonstrates by simulation the possible validity of our hypothesis and contrasts our contrastive learning results with those of supervised learning. Section IV points out to related works and Section V provides concluding remarks and outlines potential future lines of research.

## 2    Main Hypothesis

We will now present our main hypothesis and the intuition behind it. We pick handwritten digits, such as in the popular MNIST dataset (LeCun, 1998) to concretize and illustrate our discussion, but our assertions are general. Applying translation, resizing, mild rotation or mild distortion to a digit sample input (raw pixels) representation does not evidently change the perceived identity of the object, say a '4' as an example, and thus these may be considered benign transformations. In addition, it would be reasonable to assume that the distribution of the transformed samples closely resembles the distribution of the original (pre-transformed) samples. As result, the output representations of a contrastive learning model trained under such transformations will exhibit good linear separability (separability in the sequel for short). Hence, if presented with a '4', the softmax layer of a trained linear classifier (followed by hard decision) to which this contrastive learning scheme output representations are applied will generally output '0 0 0 1 0 0 0 0 0 0 ', as expected (here the classifier output vector elements represent digits in their natural order i.e. '1', '2', ….'9', '0'). Summarizing, transformation satisfies class invariance and distributions similarity, these in turn lead to separability, and, finally, separability results in good classification accuracy.

After some reflection we may be intuitively led to suspect that not only the nature of the observed object (i.e. its class) remains invariant under these transformations. Its semantically meaningful attributes do so as well. For example, for our '4' digit above each of the following attributes: "has one long quasi vertical segment", "has one short horizontal segment", "has one diagonal segment from mid-lower left to upper right" (and many more) will all remain invariant under our transformations.

This is then our main hypothesis: that transformations satisfy attributes invariance and attributes distributions similarity, that these lead to attributes separability and that attributes separability leads to accurate decisions on whether each and every attribute is present or not in the given object sample. Thus in accordance with our hypothesis, from the point of view of contrastive learning classes and their attributes will be treated alike.

We use the term 'semantically meaningful attribute' (and 'attribute' in the sequel for short) to distinguish between this term and the terms 'feature', or 'abstract feature' which are commonly used in relevant literature to describe a feature vector element, and which do not necessarily carry along with them any concrete semantically meaningful property.

To probe into the feasibility of our hypothesis we could in principle multi-label each sample in a dataset with a long list of its attributes; this would be impractically tedious. At the cost of some loss in attribute resolution we will take instead a very simple alternative approach. We notice in the above examples that each of the fore mentioned attributes may be shared by more than one class. For example "has one long quasi vertical segment" is an attribute which, in addition to '4', is shared also by '1', '6', '7' and '9'. With some abuse of language we may thus say that this attribute implies the *super-class* expression '1 0 0 1 0 1 1 0 1 0' which should be equivalently read as "our sample belongs to either class '1' or '4' or '6' or '7' or '9'", but not to '2', '3', '5', '8' nor '0'. Super-class expressions which contain a large number of 1's are implied by highly common attributes, i.e. those which are shared by many classes, and vice-versa. For example, "has curved arcs, closed curves or straight segments" is an attribute shared by all digits and thus implies an all '1's super-class expression. Composite attributes, that is such that are defined by the intersection of several 'simple' attributes (and thus contain in their semantic description several "and"s) will imply, in the limit, a single (or even no) class. An example would be the logical conjunction of the 3 fore-mentioned attributes of the digit '4'. We note that there might be some cases where 2 or more distinct attributes imply the same super-class; this would lead to the fore mentioned resolution loss.

We have $2^{10}$ super-classes (including 10 'pure' classes) in the MNIST case, each implied by one or more attributes. Following contrastive training completion and provided that our hypothesis is not false, then a trained linear classifier to which the contrastive learning output representations are fed will be able to decide with high accuracy whether a sample belongs to any selected super-class out of the $2^{10}$ super-classes, or not.

We will call this conjectured surprisingly strong capability of the output representations of contrastive learning by the term 'hyper-separability'.

We would also expect, in striking contrast, that the ubiquitous supervised trained neural network classifiers, such as (He et al., 2015), will not exhibit such hyper-separability of representations, and thus will be able to classify digit samples and infer their labels in a multi-class classification task, but not any more than that.

## 3 Probing Our Hypothesis

To probe the possible validity of our hypothesis we carry on an experiment utilizing the MNIST (LeCun, 1998; 60000 training and 10000 test samples, 10 classes, handwritten digits) and the balanced EMNIST (Cohen et al., 2017; 112800 training and 18800 test samples, 47 classes, handwritten digits and letters) datasets. MNIST allows us to check the separability of all the $2^{10}$ super-classes, while EMNIST allows only (random) super-class sampling (due to their huge quantity, $2^{47}$ in total), but provides a significantly more challenging and realistic task. Both datasets were augmented during training by means of elastic distortion (applied prior to the application of the benign transformations pair), following the suggested scheme and steps of (Simard et al., 2003).

Contrastive learning is generally agnostic to the details of the underlying neural network architecture. We implement here 2 simple fully connected feed-forward neural networks with similar architectures and 2 hidden layers each. One neural network is used for contrastive learning (784, 400, 400, 100 units per layer) and one for a standard supervised learning classifier (784, 400, 400, and 10 or 47 units per layer for MNIST and balanced EMNIST respectively). Several loss functions have been in use within the contrastive learning paradigm. We picked for this work the mini-batch based loss function which was found best for their purpose by (Chen et al., 2020). This loss, denoted $l(i, j)$, is function of $(z_i, z_j)$, the neural network last layer representations (of dimension 100 each) of the transformed pair $(x_i, x_j)$ of a given sample, and of $z_k$, the last layer representations of all other mini-batch samples $x_k$. This loss is namely expressed as (see Chen et al., 2020 for details):

$$l(i,j) = -\frac{sim(z_i, z_j)}{\tau} + \log\left(\sum_{k=1}^{2N} 1_{k \neq i} \exp\left(\frac{sim(z_i, z_k)}{\tau}\right)\right) \qquad (1)$$

where $sim(.\,,.)$ is the normalized inner product (i.e. the cosine similarity), $\tau$ is the system temperature, N is the mini-batch size, and $1_{k \neq i}$ is a binary indicator function which vanishes when k=i and equals 1 otherwise. The overall mini-batch loss will then be

$$L = \frac{1}{2N} \sum_{k=1}^{N} [l(2k-1, 2k) + l(2k, 2k-1)] \qquad (2)$$

which can be minimized by stochastic gradient descent. We have used $\tau = 1$, N = 1000 and ADAM (Kingma and Ba, 2015) optimization scheme. Again, following (Chen et al, 2020) we use the penultimate layer (and not the last layer) for output representations capture (of dimension 400) for both the contrastive learning and the supervised learning schemes (for which a standard MSE loss was used).

The benign (symmetry) transformations adopted by (Chen et al., 2020) are suitable for the Imagenet dataset (Deng et al., 2009), but not for handwritten characters.

We checked the contrastive learning multi-class classification error performance (empiric error probability, $P_{err}$) of mild elastic distortion (Simard et al., 2003), mild rotation, re-sizing and translation transformations, both separately for each transformation type (standalone) and in selected sequential transformation pairs. We found that elastic distortion alone provides the best performance. This is probably due to the fact that in both MNIST and balanced EMNIST datasets most of the characters are centered (and thus translation does not play a major role), of similar size (same for re-sizing), and of quasi-vertical position (same for rotation). We thus applied elastic distortion alone in all our experiments. Table 1 shows the multi-class test set error probability results for contrastive and supervised learning. The number of training epochs (e.g. 200 for MNIST contrastive learning) was set so that no further loss descent was observed. For both schemes, the output representations (of dimension 400), captured after training completion, were supplied as input to a multi-class supervised linear classifier with 10 (MNIST) and 47 (balanced EMNIST) units at softmax layer, which was trained for 10 epochs.

**Table 1**: Contrastive and Supervised Learning training epochs and multi-class classification error probability ($P_{err}$) for MNIST and balanced EMNIST datasets

|  | Method | Training Epochs | Test Set $P_{err}$ |
|---|---|---|---|
| **MNIST** | Contrastive | 200 | 0.014 |
|  | Supervised | 50 | 0.009 |
| **Balanced EMNIST** | Contrastive | 300 | 0.186 |
|  | Supervised | 50 | 0.132 |

It should not surprise us that, for similar contrastive and supervised learning network architectures (as those which we applied in our tests), the performance of supervised learning was significantly better than that of contrastive learning (e.g. 0.132 for supervised vs. 0.186 for contrastive, in the EMNIST case). This was also reported by (Chen et al., 2020) where in order to achieve similar error performance a wider and deeper architecture was used for contrastive learning as compared to that of supervised learning. But we are studying hyper-separability here (and not classification accuracy) so we may proceed with this.

The error performance results provided in Table 1 will serve as reference points, for checking the possible validity of our hyper-separability conjecture, as follows: multi-class classification has expected random choice error probability of 0.90 (MNIST) and 0.98 (balanced EMNIST) and thus are a much more challenging task than the 2-class super-class decision task ("does this sample belong to a specified super-class, or not?") which has, in average, 0.50 random choice error probability for both MNIST and balanced EMNIST. In order that our hyper-separability conjecture regarding contrastive learning be possibly valid we would thus expect that for contrastive learning, super-class decision error probability will be in general better than multi-class classification error probability. The opposite would be true of supervised learning: since no hyper-separability is expected in this case (there is no force driving towards super-class separability during the training period) then the 2-class super-class decision would have in general worse error performance than the more difficult multi-class classification.

Please refer to Figure 1 where histograms of super-class decision error probability are plotted for MNIST (for all its 1024 super-classes) and for balanced EMNIST (for 1000 super-classes uniformly randomly chosen from the integer interval $[0, (2^{47} - 1)]$). The multi-class classification error probabilities cited in Table 1 which serve us as references are also shown for convenience in each respective plot.

In the MNIST case, when applying contrastive learning, 93% of the super-classes exhibit ***better*** (lower) decision error probability than that of the multi-class classification task. This is in striking contrast with supervised learning where 77% of the super-classes yield ***worse*** error performance.

Balanced EMNIST shows similar results. In this case, in contrastive learning 98% of the super-classes exhibit ***better*** performance than that of the multi-class classification task, while in supervised learning 68% of the super-classes yield ***worse*** performance.

Our results above clearly support our contrastive learning related hyper-separability conjecture. Contrastive learning representations carry upon their shoulders not only class but also semantically meaningful attributes information. Supervised learning representations, in contrast, carry with them merely class information.

As mentioned above learning hyper-separable representations is a necessary but not sufficient condition to implement attribute detection. We still need a firing neuron (i.e. a separating hyperplane), which detects or decides upon each of the semantically meaningful attributes of interest (recall – we have at least $2^{47}$ attributes in the EMNIST case, only a miniscule bunch of these would be typically of interest to us). We can train such hyperplanes either supervised or unsupervised. In the supervised case we should utilize selected attributes labeled samples to train a linear classifier, as we have done for example in our experiments above. In (Nissani (Nissensohn), 2018) an unsupervised learning classifier model has been proposed. Guided by local density estimates, the model gradually shifts and rotates individual hyperplanes until they settle near the density 'valleys' which separate between (separable) classes. See (Nissani (Nissensohn), 2018) for details. Such a model, with slight modifications, can be used for our purpose here as well.

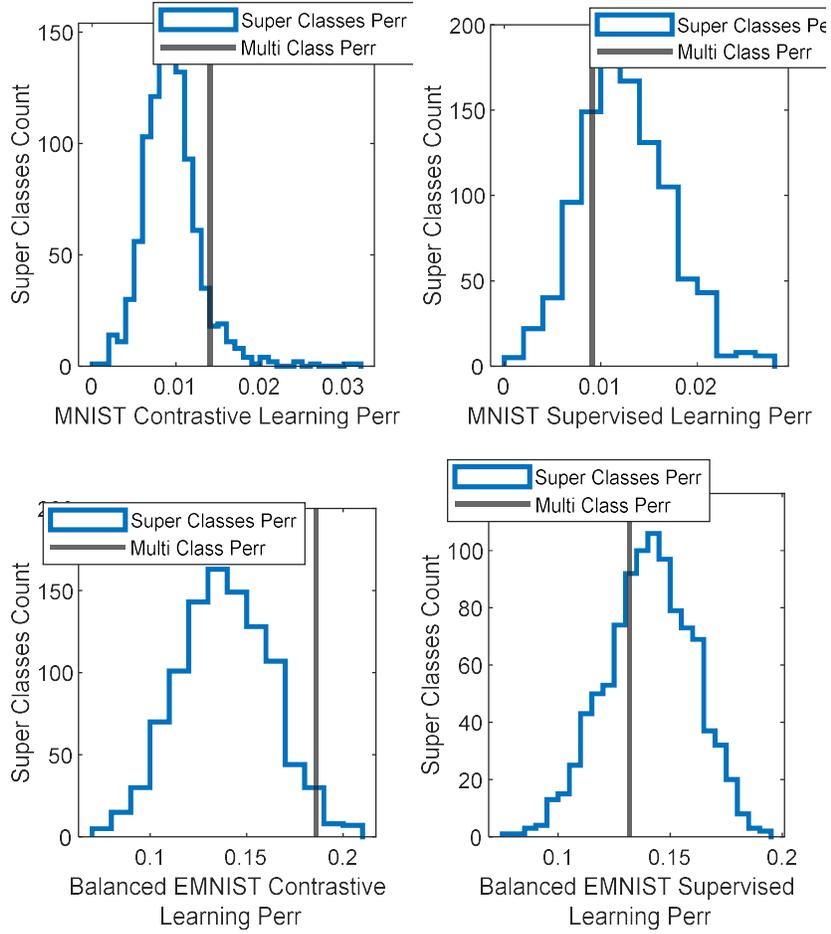

**Figure 1**: Super-class decision vs. Multi-class classification error probability.
Top: MNIST, Bottom: Balanced EMNIST. Left: Contrastive, Right: Supervised

## 4   Related Work

We are not aware of any prior neuro-computational research or study related to the spontaneous emergence of attributes associations in an unsupervised or self-supervised learning framework.

The idea that human perception results not only in an identification label but is also accompanied by a large set of associations is old and goes at least as far back as (Locke, 1689).

Connectionist models which supervised learn both extrinsic as well as intrinsic relations go back to the early days of neural networks (e.g. Rumelhart and Todd, 1993; McClelland et al., 1995) and extend up to date (Santoro et al., 2017). One of the main tracks that this approach has taken in later works is that of graph neural net models (Scarselli et al., 2009). A second similar significant track has been that of relational embeddings (Bordes et al., 2013, Wang et al., 2014) in which multi-relational knowledge bases (such as Freebase or Wordnet) are embedded in a metric space wherein the distance between the embeddings of two concepts reflects its semantic similarity. An extensive research corpus exists for both these tracks, see (Zhang et al., 2021) for a survey. Again, the proposed models of these works above are explicitly trained for their respective purposes, and this is done within the supervised learning paradigm.

The intuition lying behind contrastive learning is that identity preserving transformations are naturally available to living organisms (through variation of object pose and lighting, through object manipulation, and so on) and that these, complemented by a (possibly innate) invariance principle prior, can become an important instrument to arrive to useful internal representations. This intuition has guided many researchers, including (Hadsell et al., 2006; Bouvrie et al., 2009; diCarlo et al., 2012; Anselmi et al., 2013).

We have followed in general (Chen et al., 2020) contrastive learning scheme due to its methodical and slim approach. We have replaced their ResNet (He et al., 2015) neural network of choice by a much simpler one (we are not competing on accuracy performance in this work and our datasets are less demanding) and replaced their set of benign transformations (suitable for the Imagenet dataset) with our own. One possible disadvantage of this scheme is its computational and memory complexity for mini-batch processing and storage as reflected in Equations (1) and (2) above. There exists a significant line of work on contrastive learning and we could possibly have similarly picked any of e.g. (Grill et al., 2020; Chen and He, 2020; Mikolov et al., 2013; Dosovitskiy et al., 2014; Oord et al., 2018; Bachman et al., 2019).

## 5    Concluding Remarks

We have provided demonstrative evidence that the output representations of a contrastive learning scheme carry along with them information not only about class identity but also regarding attribute relations of the perceived object. This property which we have denoted 'hyper-separability' is characteristic of contrastive learning and does not seem to show up in supervised learning schemes.

Interestingly, this identity and relational information appears in a single representation package, as a single "bundle", as opposed for example to some neural graph models where object identity and attribute relations are represented as separate entities (nodes and edges). In order to express this bundle contents as a semantically meaningful sentence it would be required to sequentially fetch the information of interest, piece by piece (Bickerton, 1990). To do this one should first extract (decode) this embedded information. A simple way to carry out such information extraction consists, as was described above, in placing linear multi-class classifiers (for identity) and 2-class classifiers (for attributes of interest) as a last 'decoding' neural layer.

Significant efforts (Zeiler and Fergus, 2014; Yosinski, 2015; Nguyen et al., 2016) have been invested in attempts to decipher the attributes or interpretable features which individual neurons of neural nets hidden layers maximally respond to. These studies have been carried solely upon supervised learning networks. After reading the present work, it should not be a major surprise to us that they all resulted in no major findings. It seems that attributes related information not only is available in contrastive learning but not in supervised learning, but also that, perhaps counter-intuitively, it is available in the output representation layer and not embedded within hidden layers; the attribute responsive neurons are output neurons, just like classification neurons, and they are fed by the same output representation layer that feeds the final linear classification layer. This is a more 'democratic' view than the traditional one where attributes have been traditionally viewed as lower level components of a hierarchical structure. In this respect this paper apparent observation and discovery may also result in the Gordian knot cutting of the infamous 'Binding Problem' (Riesenhuber and Poggio, 1999; von der Malsburg, 1981).

It has been suspected for decades that animal learning is mainly based upon some unsupervised (or self-supervised) strategy. The works of (Chen et al., 2020) and others which demonstrate contrastive learning competitive accuracy performance along with this present work (spontaneous attributes association) may provide additional support for the idea that benign transformations combined with a tacit innate principle of permanence or continuity (which are the main components of contrastive learning) constitute one of the basic strategies for animal learning.

We have dealt in this work with the emergence of attributes related associations, namely intrinsic relations. The seamless emergence of environment related associations, i.e. extrinsic relations, remains an open problem, and is left for future work. Perhaps answers to this could be found within the Reinforcement Learning (Sutton and Barto, 2018) paradigm, which purposely deals with the interaction between organisms and their surrounding environment.

To facilitate replication of our results a simulation package will be provided by the author upon request.